\pdfoutput=1

\documentclass[11pt]{article}

\usepackage[final]{acl}
\PassOptionsToPackage{most}{tcolorbox}
\usepackage{tcolorbox} 

\usepackage{times}
\usepackage{latexsym}
\usepackage{placeins} 
\usepackage{tcolorbox}
\usepackage{enumitem}
\usepackage{tcolorbox} 
\usepackage[T1]{fontenc}

\usepackage[utf8]{inputenc}
\usepackage{tcolorbox}
\usepackage{enumitem}
\usepackage{microtype}

\definecolor{cadmiumgreen}{RGB}{0,107,60}
\usepackage{csquotes}  
\usepackage{natbib}    
\usepackage{amsmath, amssymb}
\usepackage{marvosym}
\newcommand\blfootnote[1]{%
\begingroup
\renewcommand\thefootnote{}\footnote{#1}%
\addtocounter{footnote}{-1}%
\endgroup
}

\usepackage{algorithm}  
\usepackage{multirow} 
\usepackage{makecell}
\usepackage{natbib}
\usepackage{algpseudocode}  
\usepackage{graphicx}
\usepackage{csquotes}
\usepackage{booktabs}
\usepackage{algorithm}
\usepackage{xcolor}
\usepackage{listings}
\usepackage[most]{tcolorbox}
\newtcolorbox{dialogbox}{
  colback=gray!10!white, colframe=black, sharp corners,
  boxrule=0.5mm, top=10pt, bottom=10pt, left=10pt, right=10pt, breakable
}
\lstdefinestyle{smallsqlstyle}{language=SQL,basicstyle=\scriptsize\ttfamily,keywordstyle=\color{blue},keepspaces=true}
\lstdefinestyle{sqlstyle}{language=SQL,keywordstyle=\color{blue},keepspaces=true}
\lstdefinestyle{schemastyle}{language={},columns=flexible}

\newcommand{\posi}{\includegraphics[height=0.8em]{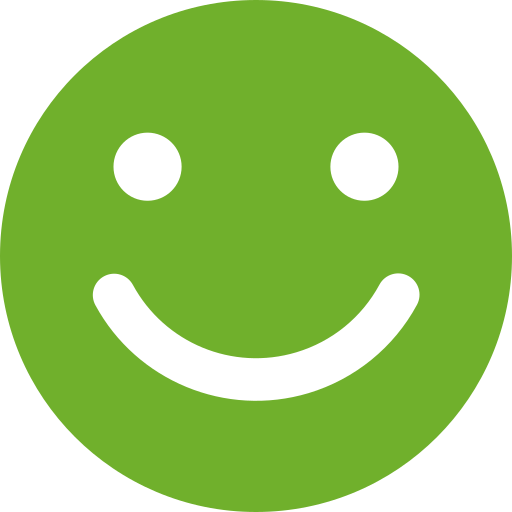}}
\newcommand{\nega}{\includegraphics[height=0.8em]{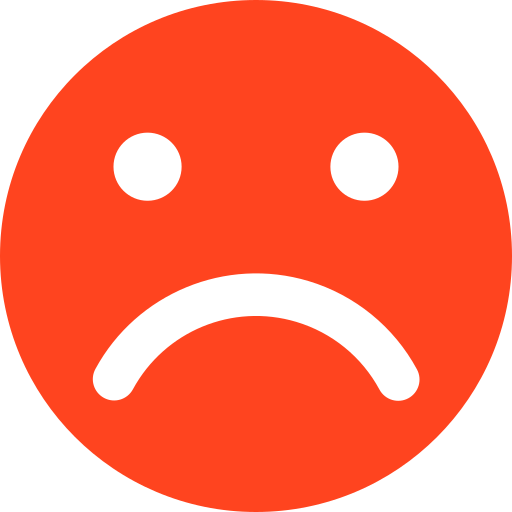}}
\newcommand{\normal}{\includegraphics[height=0.8em]{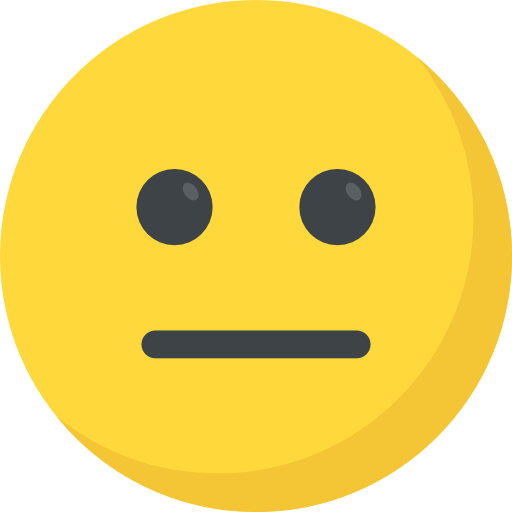}}

\lstdefinestyle{smallsqlstyle}{language=SQL,basicstyle=\scriptsize\ttfamily,keywordstyle=\color{blue},keepspaces=true}
\lstdefinestyle{sqlstyle}{language=SQL,keywordstyle=\color{blue},keepspaces=true}
\lstdefinestyle{schemastyle}{language={},columns=flexible}

\usepackage{inconsolata}

\usepackage{graphicx}

\title{Toward Structured Knowledge Reasoning:\\ Contrastive Retrieval-Augmented Generation on Experience}

\author{
\textbf{Jiawei Gu\textsuperscript{1,2}},
\textbf{Ziting Xian\textsuperscript{1}},
\textbf{Yuanzhen Xie\textsuperscript{2}},
\textbf{Ye Liu\textsuperscript{2}},
\\
\textbf{Enjie Liu\textsuperscript{2}},
\textbf{Ruichao Zhong\textsuperscript{2}},
\textbf{Mochi Gao\textsuperscript{2}},
\textbf{Yunzhi Tan\textsuperscript{2,$\dag$}},
\textbf{Bo Hu\textsuperscript{2,$\dag$}},
\textbf{Zang Li\textsuperscript{2}}
\\
\textsuperscript{1}Sun Yat-sen University
\textsuperscript{2}Platform and Content Group, Tencent
}

\begin{document}
\maketitle
\blfootnote{$\dag$ \ Corresponding author. For further information or academic correspondence regarding this work, please contact: \Letter\ kuvvius@gmail.com, \{boristan, bohu\}@tencent.com.}

\begin{abstract}
Large language models (LLMs) achieve strong performance on plain text tasks but underperform on structured data like tables and databases. Potential challenges arise from their underexposure during pre-training and rigid text-to-structure transfer mechanisms.
Unlike humans who seamlessly apply learned patterns across data modalities, LLMs struggle to infer implicit relationships embedded in tabular formats, especially in the absence of explicit structural guidance.
To bridge this cognitive gap, we introduce \underline{\textbf{Co}}ntrastive \underline{\textbf{R}}etrieval-Augmented Generation on \underline{\textbf{E}}xperience~(\underline{\textbf{Co}}\underline{\textbf{R}}\underline{\textbf{E}}), a framework that builds experience memory representations and enhances generalization through contrastive in-context learning (ICL) to simulate human-like knowledge transfer. Experiments on Text-to-SQL and TableQA show CoRE significantly improves performance, achieving average gains of 3.44\% and 4.24\%, with up to 17.2\% on challenging tasks. Our Monte Carlo Tree Search (MCTS)-generated Experience Memory expands training data 8-9$\times$, enhancing diversity and domain coverage. This training-free and continual method propels LLMs toward structured knowledge expertise. Code is available at: \url{https://github.com/Kuvvius/CoRE}.

\end{abstract}

\section{Introduction}

Large language models (LLMs) have shown remarkable capabilities on plain text but struggle with low-frequency, domain-specific concepts~\cite{ domain_rag_survey, gu2024cmr} and structured data~\cite{structlm, structgpt, table_meets_llm} like tables and databases. 
This arises from two key issues: (1) limited exposure of LLMs to structural relationships during pre-training~\cite{longtail_survey, xu2024chartmoe}, and (2) difficulty in transferring reasoning from text to structured data~\cite{olagpt, dater}. These issues are especially evident when reasoning with structured knowledge, where understanding the connections between structured data pieces and query text is crucial for accurate conclusions.

\begin{figure*}[t]
  \vspace{-15pt}
  \centering
  \includegraphics[width=2\columnwidth]{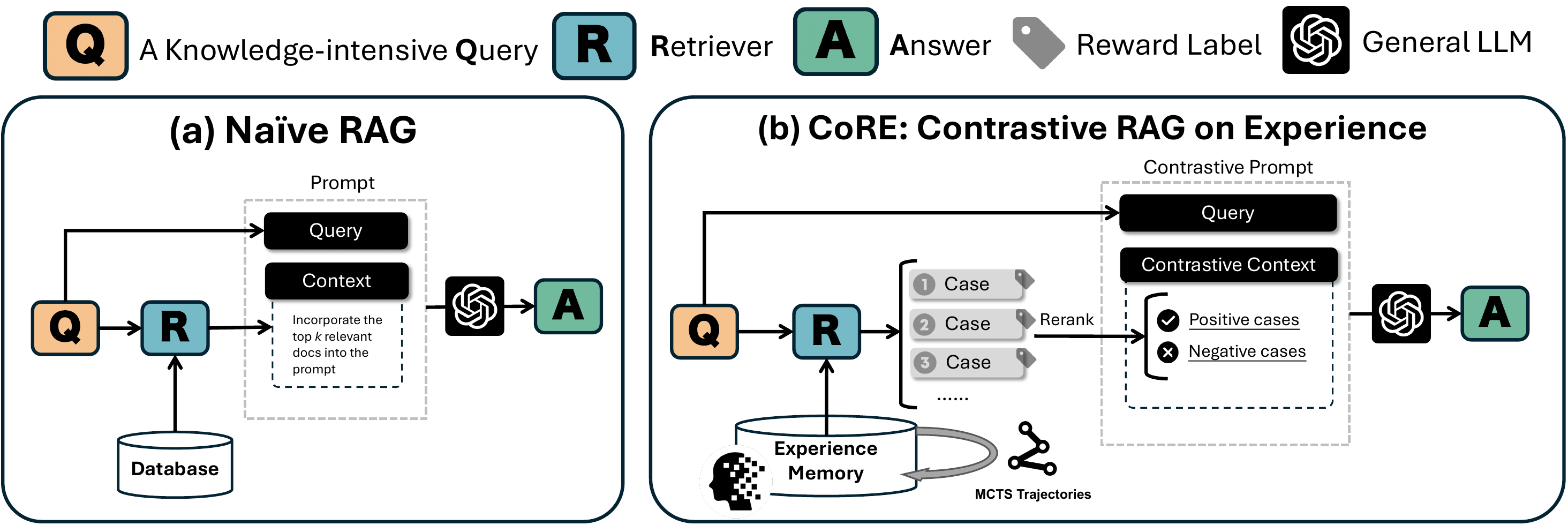}
  \caption{Comparison between naive RAG and our CoRE.}
  \label{fig:fig1_rag_vs_core}
  \vspace{-15pt}
\end{figure*}

To reduce reliance on pre-training data, Retrieval-Augmented Generation (RAG)~\cite{rag_knowledge_intensive_2020, gao2023retrieval} has become a widely adopted technique for injecting external knowledge into LLMs. While effective in text-based tasks~\cite{domain_rag_survey}, naïve RAG shown in Figure~\ref{fig:fig1_rag_vs_core}~(a), fails to improve performance due to the lack of specialized mechanisms for structured knowledge reasoning. Adding disconnected structured components (e.g. schemas and sub-tables) without grasping their interrelationships can disrupt the reasoning chain and hinder performance~\cite{mac-sql,dail-sql}.

While RAG struggles with structured knowledge reasoning, humans excel at leveraging and transferring past experiences to solve complex, specialized problems, even those they have less exposure to. According to \textbf{\textit{Deliberate Practice}}~\cite{peak_deliberate_learning}, human experts deliberately expose themselves to diverse schema cases, building mental representations that compensate for initial knowledge gaps~(more explanation in Appendix~\ref{MR}).
This enables them to solve unfamiliar problems by reflecting on both successful and failed attempts (e.g., joining database tables correctly or incorrectly). Through self-monitoring, they identify mistakes, refine strategies, and focus on transferring patterns to structural reasoning rather than just memorizing surface-level facts.

To adapt this human-like learning process to LLMs, the focus shifts to building dynamic experience and leveraging contrastive reflection. Previous works on contrastive learning~\cite{contrasive_learning}, alignment~\cite{rlhf, prm800k}, and in-context learning~\cite{contrastive_cot} have shown that LLMs can learn more effectively by both positive and negative examples. This idea has been further supported by studies on reasoning~\cite{agentq}, which improve the generalization of contrastive trajectories in complex, multi-step reasoning tasks.

Therefore, we introduce \underline{\textbf{Co}}ntrastive \underline{\textbf{R}}etrieval-Augmented Generation on \underline{\textbf{E}}xperience~(\underline{\textbf{Co}}\underline{\textbf{R}}\underline{\textbf{E}}). \textit{\textbf{CoRE builds experience memory representations and incorporates contrasted examples}}—both successful and failed cases—\textit{\textbf{into the context}} to solve structured knowledge reasoning problems. 
As shown in Figure~\ref{fig:fig1_rag_vs_core}~(b), CoRE simplifies intricate structured reasoning problems into decomposable ones. It employs Monte Carlo Tree Search (MCTS)~\cite{silver2017mastering, RAP} along with a self-evaluation mechanism~\cite{llm-as-a-judge} to mimic human \textit{Deliberate Practice}, heuristically guiding the collection of successful and failed trajectories. Through \textit{selection}, \textit{expansion}, \textit{simulation}, and \textit{back-propagation}, question-answer pairs are generated with reward labels (representing the probability of success in intermediate steps). Subsequently, when a new problem arises, CoRE retrieves relevant cases and re-ranks them based on reward labels to present both successful~(positive) and failed~(negative) examples for contrastive reflection, enabling more accurate and knowledge-intensive answers.

Extensive experiments on structured knowledge reasoning tasks, such as Text-to-SQL and TableQA, show that CoRE significantly enhances both performance and efficiency, yielding average gains of 3.44\% and 4.24\% over baseline methods. In particular, CoRE achieves greater improvements on more challenging problems, with increases of 17.2\% and 8.2\% (Figure~\ref{fig:fig4_difficulty_bird}), respectively. 
The Experience Memory Dataset built from MCTS trajectories expands the original training dataset by 8-9$\times$, demonstrating increased diversity and domain coverage with diluted, decomposed high-density knowledge (Figure~\ref{fig:fig3_reward_distribution}). 
Moreover, compared to few-shot approaches using only fixed examples, CoRE underscores the effectiveness of contrastive learning in context, achieving a 1.76\% gain (Table~\ref{tab:ablation_study_2_shot}). In summary, our main contributions are as follows:
\vspace{-15pt}
\begin{enumerate}
    \item \textbf{Experience Memory Construction}: We introduce a method to build experience memory using MCTS trajectories, capturing diverse reasoning experiences (both successful and failed), enabling better generalization in complex tasks with minimal initial exposure.\vspace{-5pt}
    \item \textbf{Contrastive Reflection Mechanism}: CoRE introduces a contrastive learning method, which obtains two ranks through re-ranking in the RAG process and then incorporating retrieved content with specialized mechanisms for structured knowledge reasoning.\vspace{-5pt}
    \item \textbf{Scalable Knowledge Retrieval}: CoRE offers a plug-and-play, training-free solution, allowing for continuous learning and retrieval in evolving environments by dynamic updates to the Experience Memory.
\end{enumerate}
\vspace{-10pt}

\section{Preliminary}
\label{sec:preliminary}

\vspace{-3pt}

\noindent In this section, we present our main Research Questions and the formulation of \textbf{Structured Knowledge Reasoning}, which uses LLMs to solve complex reasoning tasks based on structured data.

\vspace{-2pt}
\begin{tcolorbox}[colback=lightgray!10, colframe=black, title={Research Questions}]
(1)~Can MCTS effectively imitate human's practice for heuristic data augmentation? 
(2)~Can RAG enhance structured knowledge reasoning? 
(3)~Does contrastive ICL outperform fixed examples?
\end{tcolorbox}
\vspace{-2pt}

\noindent Existing methods for structured knowledge reasoning seldom employ RAG because retrievable, domain-specific data is scarce. Instead, retrieval from general documents often introduces noise, leading to a negative impact~\cite{corrective_rag,liu2024lost}.
For instance, DAIL-SQL~\cite{dail-sql} restricts retrieval to training data, limiting diversity and coverage. Second, agent-based methods~\cite{mac-sql,structgpt,din-sql} rely on fixed few-shot examples without incorporating dynamic contrastive learning for human-like adaptation. Third, most approaches lack systematic mechanisms to evaluate positive/negative signal impacts during reasoning. These gaps drive our investigation into MCTS-based augmentation, RAG optimization, and contrastive learning for structured knowledge reasoning with LLMs.

\vspace{-3pt}
\paragraph{Formalization}
Structured knowledge reasoning can be described as a question-answering task. 
Given a triple $\mathcal{X} = (\mathcal{Q}, \mathcal{K}, \mathcal{A})$, where $\mathcal{Q}$ represents a natural language question, $\mathcal{K}$ is a structured knowledge representation, and $\mathcal{A}$ is an evidence-supported answer. Based on the natural language question $\mathcal{Q}$ and the accessible structured data $\mathcal{K}$ (e.g., a table or database), the first step is to extract useful evidence $\mathcal{E}$ from $\mathcal{K}$ (i.e., structured knowledge grounding). 
Notably, $\mathcal{E}$ remains a structured, high-density knowledge representation. Then, the LLM generator $\mathcal{G}$ uses this evidence to generate the expected answer $\mathcal{A}$ in response to the question $\mathcal{Q}$. According to the task requirement, the generated answer $\mathcal{A}$ must be evidence-supported, but either free-form answers in natural language or structured expressions (e.g. SQL statements) to be executed for obtaining the answer from $\mathcal{K}$. This process can be formalized as:
\begin{equation}
\mathcal{A} \overset{\text{\tiny reasoning}}{\sim} \mathcal{G}\left(\mathcal{Q}, \left( \mathcal{E} \overset{\text{\tiny grounding}}{\leftarrow} (\mathcal{Q}, \mathcal{K}) \right) \right)
\end{equation}
\vspace{-3pt}
In our work, \textbf{we focus on improving reasoning over extracted structured evidence, excluding the grounding phase}. To better illustrate our task, we detail two types of reasoning formulation on databases and data tables:

\vspace{5pt}
\noindent \textbf{Text-to-SQL  } We define a database \(\mathcal{K} = \{\mathcal{T}, \mathcal{C}\}\), where \(\mathcal{T} = \{t_1, t_2, \dots, t_{|\mathcal{T}|}\}\) denotes tables and \(\mathcal{C} = \{c_1, c_2, \dots, c_{|\mathcal{C}|}\}\) denotes columns, with foreign keys \(\{(c_i^{(k)}, c_j^{(h)})\}\) linking them; consequently, the grounding phase selects from \(\mathcal{K}\) the relevant tables and columns for \(\mathcal{Q}\), forming evidence \(\mathcal{E} = \{(t_k, c_i) \mid t_k \in \mathcal{T}_{\mathcal{E}},\, c_i \in \mathcal{C}_{\mathcal{E}}\}\) that the generator then uses to produce the answer \(\mathcal{A}\).
The reasoning process employs $\mathcal{G}$ to derive the final answer by sampling from its generative distribution. For more specific examples, refer to Appendix~\ref{apd:prompt_Details}.

\vspace{5pt}
\noindent \textbf{TableQA  } We model a data table \(\mathcal{K} = \{\mathcal{C}, \mathcal{R}\}\), where \(\mathcal{C} = \{c_1, c_2, \dots, c_{|\mathcal{C}|}\}\) are columns and \(\mathcal{R} = \{r_1, r_2, \dots, r_{|\mathcal{R}|}\}\) are rows, with each cell value denoted by \(v_{i,j}\); accordingly, the grounding process identifies from \(\mathcal{K}\) the pertinent columns \(\mathcal{C}_{\mathcal{E}}\) and rows \(\mathcal{R}_{\mathcal{E}}\) based on \(\mathcal{Q}\), forming evidence \(\mathcal{E} = \{v_{i,j} \mid c_i \in \mathcal{C}_{\mathcal{E}},\, r_j \in \mathcal{R}_{\mathcal{E}}\}\) that supports $\mathcal{G}$ deriving \(\mathcal{A}\).

\section{CoRE}
\begin{figure*}[t!]
  \includegraphics[width=2\columnwidth]{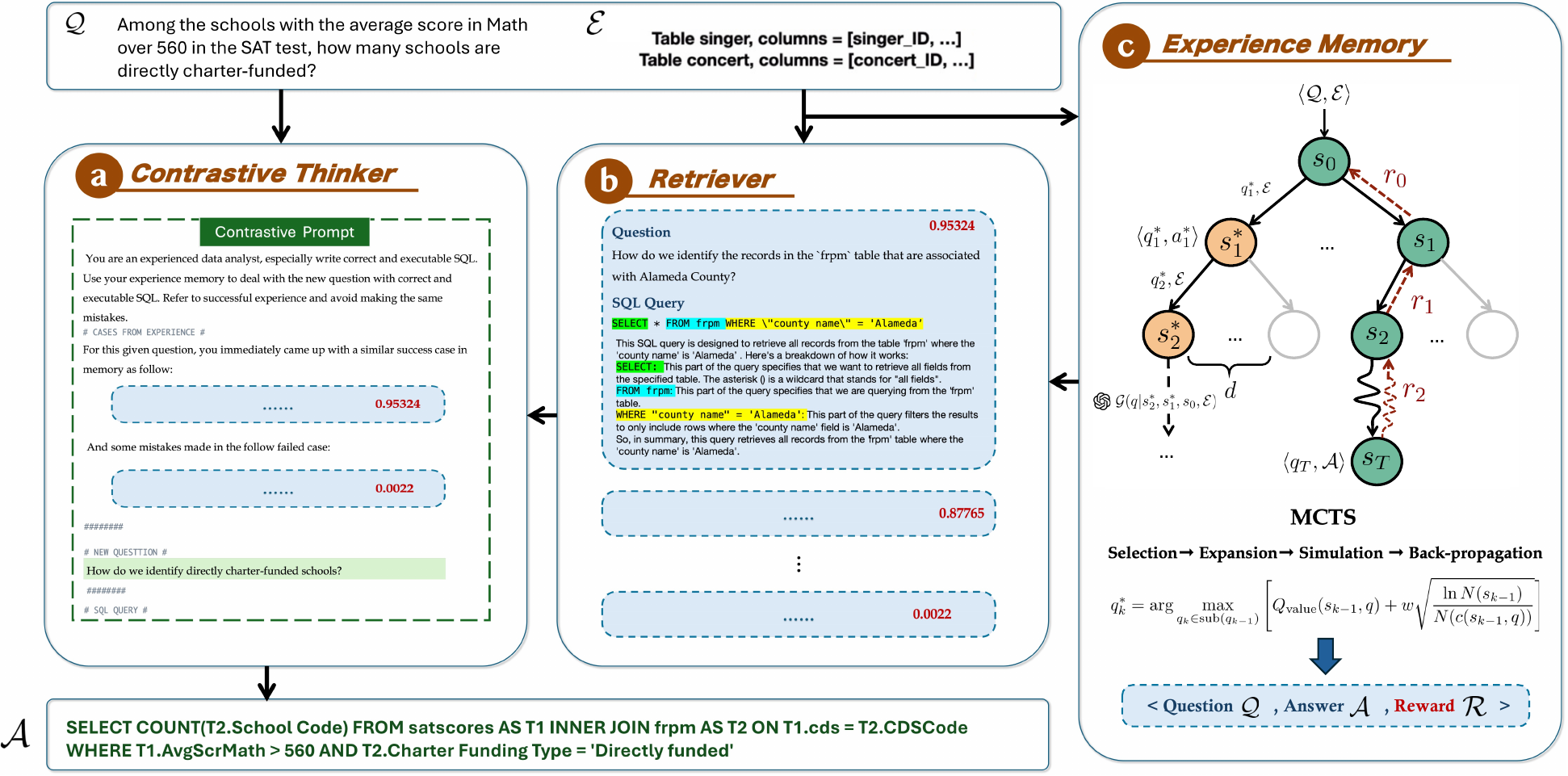}
  \caption{Overall framework of CoRE.}
  \label{fig:fig2_core_framework}
  \vspace{-8pt}
\end{figure*}

We present our \underline{CoRE} framework in this section. As illustrated in Figure~\ref{fig:fig2_core_framework}, it consists of three key modules: \underline{{Co}}ntrastive Thinker~(\S~\ref{method:contrastive_thinker}), \underline{{R}}etriever~(\S~\ref{method:retriever}) and \underline{{E}}xperience Memory Buildier~(\S~\ref{method:experience_memory}).

\subsection{Experience Memory Builder}\label{method:experience_memory}
To tackle complex reasoning tasks involving questions $\mathcal{Q}$ with structured knowledge $\mathcal{E}$, humans and agents decompose problems into simpler sub-questions. Each sub-question and answer pair, $\langle q, a \rangle$, represents an intermediate stage in a reasoning trajectory $\tau = \{\langle q_1, a_1 \rangle, \langle q_2, a_2 \rangle, \dots, \langle q_n, a_n \rangle\}$, leading to the final answer $\mathcal{A}$. This decomposition transforms reasoning into a planning problem, where the goal is to find the optimal trajectory:
$
    \tau^* = \{\langle q_1^*, a_1^* \rangle, \langle q_2^*, a_2^* \rangle, \dots, \langle q_n^*, a_n^* \rangle\}.
$
We employ MCTS with balanced exploration and exploitation to construct trajectories (see Figure~\ref{fig:fig2_core_framework}(c) and Algorithm~\ref{algo:tree-based_search}) involving four phases:

\paragraph{\textbf{Selection.}}  
Starting from the root node (representing $\mathcal{Q}$ and $\mathcal{E}$), the algorithm recursively selects child nodes based on the Upper Confidence Bound for Trees (UCT):


{
\vspace{-10pt}
\small
\begin{equation}
\label{eq:uct}
\begin{aligned}
    q_k^\ast = & \arg\max_{q_k \in \text{sub}(q_{k-1})} \quad  \\
    & \left[
        Q_\text{value}(s_{k-1}, q) 
        + w \sqrt{\frac{\ln N(s_{k-1})}{N(c(s_{k-1}, q))}}
    \right],
\end{aligned}
\end{equation}
}

\vspace{-3pt}
\noindent where$s_{k-1} = \langle q_{k-1}, a_{k-1} \rangle$, $N(s_{k-1})$ is the visit count for node $s_{k-1}$, and $c(s_{k-1}, q_k)$ is the corresponding child node. The weight $w$ tunes the balance between exploration (less visited nodes) and exploitation (high $Q_\text{value}$ nodes).

\vspace{3pt}
\noindent \textbf{Expansion.}
Upon reaching a leaf node, we generate $d$ candidate sub-questions using $\mathcal{G}$:
$\text{sub}(q_k) \sim \mathcal{G}(q \mid s_{k-1}, \dots, s_0, \mathcal{E}).$
If the leaf is terminal (i.e., the end of a reasoning chain), expansion is skipped.

\noindent \textbf{Simulation  }
To estimate the expected future rewards ($Q_\text{value}$), this phase simulates future scenarios of the current node using the world model. Starting from the current node, at each state $s$, a sub-question is generated according to a \emph{roll-out policy}, and the world model predicts the next state. This roll-out continues until a terminal state is reached. In CoRE, to simplify the process and reduce noise, we adopt a similar approach as in the expansion phase, generating $d$ candidate sub-questions and selecting the one with the highest local reward, $q_k^* = \max_{q_k} f_r(s_{k-1}, q_{k})$. For practical efficiency, we use a simplified reward function $f_r$ to select sub-questions during simulation: $f_r(s_{k-1}, q_k) = r_1^\alpha \cdot r_2^{1-\alpha}$,
where $r_1$ is the consistency reward for $q_k$, calculated by counting the frequency of the $d$ sub-questions $\text{sub}(q_{k-1})$, and $r_2$ is the self-evaluation reward from the generator, defined as $r_2 = \mathcal{G}(\text{"Yes"} \mid q_k, \text{prompt})$, with $\text{prompt} = \texttt{"Is this reasoning step correct?"}$.

\noindent \textbf{Back-propagation}
Upon reaching a terminal state $s_T$ (i.e., determining the final answer) in the previous phases, we obtain a reasoning path from the root node to the terminal node. At this point, we back-propagate the rewards along the path to update the $Q_\text{value}$ for each question-answer pair. Specifically, $Q_\text{value}(s_{k-1}, q_{k})$ is updated by aggregating the rewards from all future steps of node $s_{k}$:
\begin{align}
    Q_\text{value}(s_{k-1}, q_k) 
    &= \max_{s_{k-1}, q_k, r_k, \dots, s_l, q_l, r_l, s_{l+1}, \dots} \nonumber\\
    &\quad \operatorname{avg}(r_{k-1}, \dots, r_l, \dots).
    \label{eq:q-avg}
\end{align}
After completing the predetermined MCTS iterations, the algorithm terminates and extracts the final reasoning trace by starting at the root and iteratively selecting the action with the highest $Q_\text{value}$ until a terminal node is reached.

\vspace{-3pt}
\subsection{Retriever}\label{method:retriever}
The Retriever in the CoRE framework identifies both positive and negative cases for the new question \( \mathcal{Q}_{\text{current}} \). As showed in Figure~\ref{fig:fig2_core_framework}~(b), it searches the Experience Memory database \( \mathcal{D}_{\text{experience}} \) by measuring semantic similarity \( \text{Sim}(\mathcal{Q}_{\text{current}}, \mathcal{Q}_{e_i}) \) between \( \mathcal{Q}_{\text{current}} \) and stored experience entries \( \{e_1, e_2, \dots, e_k\} \). The top \( k \) entries are retrieved and ranked by similarity, denoted as \( \text{rank\_sim}(e_i) \). These entries, which include structured knowledge and natural language descriptions, enhance retrieval accuracy. For example, SQL queries are accompanied by natural language descriptions of their Abstract Syntax Trees (AST)~(Example in the Figure~\ref{fig:fig2_core_framework}), improving the semantic understanding of retrieval and aligning the entries more closely with the query semantics.
Next, the top \( k \) entries are re-ranked based on their reward labels \( r(e_i) \), producing \( \text{rank\_reward}(e_i) \). The final ranking combines both similarity and reward ranks: \( \text{rank\_final}^{\text{positive}}(e_i) \) merges \( \text{rank\_sim}(e_i) \) and \( \text{rank\_reward}(e_i) \), while \( \text{rank\_final}^{\text{negative}}(e_i) \) merges \( \text{rank\_sim}(e_i) \) with the reverse of \( \text{rank\_reward}(e_i) \). The final ranking is obtained by averaging these components. Finally, based on the n-shot setting, the top \( n \) positive and negative entries are selected for the next stage.

\begin{figure*}[h]
  \vspace{3pt}
  \centering
  \includegraphics[width=1.8\columnwidth]{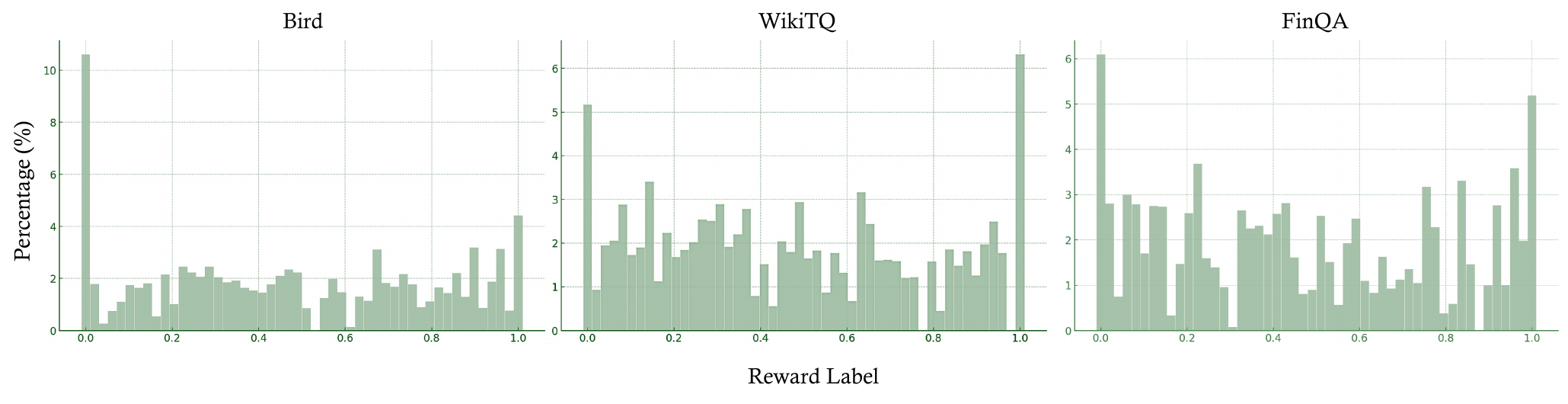}
    \caption{Reward label distribution on \textit{Bird}, \textit{WikiTQ} and \textit{FinQA}}
  \label{fig:fig3_reward_distribution}
  \vspace{-18pt}
\end{figure*}

\vspace{-5pt}
\subsection{Contrastive Thinker}\label{method:contrastive_thinker}

The Contrastive Thinker is built upon a general LLM that is specifically prompted to generate the necessary text based on the given contrastive context. Upon receiving positive and negative cases from the Retriever, the Contrastive Thinker employs deliberate practice by treating positive cases as past successes and negative cases as past failures. Through In-Context Learning, it can monitor its own performance, identify mistakes, and make adjustments accordingly. This process is encapsulated in the directive: \texttt{\small“Solve the \textless\textless\textless NEW CASE \textgreater\textgreater\textgreater using the successful experience and avoid repeating the errors from failed experience...”}, as illustrated in Figure~\ref{fig:fig2_core_framework}~(a). 
The examples of prompts constructed through the collaboration of Retriever and Contrastive Thinker can be found in the Appendix~\ref{apd:prompt_Details}.

\vspace{-4pt}
\section{Experiment}
\vspace{-4pt}
In this section, we first outline the experiment setup in \S~\ref{Exp:experiment_setup} and describe the Experience Memory dataset we constructed in \S~\ref{Exp:experience_memory_dataset}. Then, we present experiments and results for two structured knowledge reasoning tasks in \S\ref{Exp:result_and_analysis}, including TableQA and Text-to-SQL. Finally, the ablation study~(\S~\ref{Exp:ablation_study}) and strategy analysis~(\S~\ref{Exp:strategy_and_discussion}) are presented.

\vspace{-3pt}
\subsection{Experiment setup}~\label{Exp:experiment_setup}
\noindent We use LlaMA-3 series~\cite{llama3} and OpenAI's GPT-4~\cite{gpt4} as feneral LLMs in different experiment settings. The embedding models we used for retrieval is \texttt{bge-large-en-v1.5}. More experiment details in Appendix~\ref{app:exp_details}.

\paragraph{Datasets.}
We evaluate our CoRE on the dev sets on the training sets of different structured knowledge reasoning datasets showed in Table~\ref{tab:dataset_stats}, including  WikiTQ~\citep{wiki_table_question}, FinQA~\citep{finqa} and Bird~\cite{bird}. The details of these two datasets are provided in Appendix~\ref{apd:datasets}.

\paragraph{Evaluation.}  
For Text-to-SQL, we use Execution Accuracy (EX), which measures the correctness of the SQL execution, to evaluate our method on Bird~\cite{bird}. For TableQA, accuracy is used to verify whether predicted answers match the gold ones on WikiTQ~\cite{wiki_table_question} and FinQA~\cite{finqa}.

\paragraph{Baselines.}
CoRE is a plug-and-play framework, so we selected prompt-based agent reasoning frameworks as baselines~\cite{din-sql,dail-sql,mac-sql,structgpt,dater}. We integrate CoRE into their components or replace their few-shot approaches. Note that we have not altered the grounding method but only integrated CoRE into the reasoning component. A detailed introduction to various methods is provided in Appendix~\ref{apd:baselines}, and differences between their approaches and ours are discussed below.

\subsection{Experience Memory Dataset}\label{Exp:experience_memory_dataset}
The method we used to build the Experience Memory Dataset is detailed in \S~\ref{method:experience_memory} and Algorithm~\ref{algo:tree-based_search}. For cost considerations, we used LlaMA-3-70b as backbone to produce experience memory data. During the MCTS process to imitate human thinking, we set the capacity of candidate sub-questions $K=4$, roll-out times in simulation $N=10$, depth limit $L=5$, the balance weight $w = 0.5$, and reward weight $\alpha = 0.5$. To encourage more diverse reasoning paths, we set the sampling temperature to $0.8$. On average, each structure knowledge reasoning question can be decomposed into 3-5 $\langle q, a \rangle$ pairs in each iteration.
To obtain more reliable reward labels for each $\langle q, a \rangle$, in the search tree we built after 10 rolls-outs, the $Q_\text{value}$of the nodes with more than or equal to 3 visits is stored as the reward label of $\langle q, a ,r\rangle$. Moreover, we add overall successful cases with reward labels "$1.00$" and failed cases with "$0.00$" to the dataset as experience.

Finally, we construct the Experience Memory Dataset on the training set (Table~\ref{tab:dataset_stats}), expanding the original data by 8-9$\times$. The reward label distribution is shown in Figure~\ref{fig:fig3_reward_distribution}, demonstrating \textbf{increased diversity and domain coverage} with diluted and decomposed high-density knowledge~(Figure~\ref{fig:fig3_reward_distribution}). The $\langle q, a ,r\rangle$ with $r \in (0,1)$ is not included in the original training dataset, and they have difficulty difference after decomposition. In cases of errors, data entries with lower reward values usually come from nodes close to the error endpoint. On the other hand, data entries with higher rewards are more likely to lead to the correct endpoint after undergoing multiple verifications.

\vspace{-5pt}
\subsection{Result and Analysis}\label{Exp:result_and_analysis}

{
\vspace{-3pt}
\footnotesize
\begin{table}[h!]
    \caption{Results by LlaMA3-70b or GPT-4, evaluated by Execution Accuracy~(EX) on Bird Dev. }
    \label{tab:bird_results}
    \centering
    \small
    \begin{tabular}{l|c}
      \toprule
       \textbf{Method} & \textbf{EX~(\%)} \\
      \midrule
         DIN-SQL~\cite{din-sql}       &  30.5 \\
       \multicolumn{1}{r|}{\hspace{1cm}+ our CoRE}       &  34.0~($\uparrow$ 3.5) \\
       \midrule
       DAIL-SQL~\cite{dail-sql}      &  31.6\\
       \multicolumn{1}{r|}{\hspace{1cm}+ our CoRE}       &  35.2~($\uparrow$ 3.6) \\
       \midrule
       DAIL-SQL~(SC)~\cite{dail-sql}    &  37.5 \\
       \multicolumn{1}{r|}{\hspace{1cm}+ our CoRE}       &  36.7~($\downarrow$ 0.8) \\
       \midrule
       MAC-SQL~\cite{mac-sql}       &  34.9 \\
       \multicolumn{1}{r|}{\hspace{1cm}+ our CoRE}       &  40.8~($\uparrow$ 5.9)  \\
       \midrule
       MAC-SQL~(GPT-4)~\cite{mac-sql}       &  46.6 \\
       \multicolumn{1}{r|}{\hspace{1cm}+ our CoRE}       &  51.6~($\uparrow$ 5.0) \\
      \midrule
       \textbf{$\Delta$ CoRE}  & 3.44 \\
      \bottomrule
    \end{tabular}
    \vspace{-8pt}
\end{table}
}

\paragraph{Text-to-SQL.}\label{result:text2sql}
Table~\ref{tab:bird_results} shows overall performance of all methods. 
To ensure consistency with CoRE, baselines were run in a 2-shot setting. On average, CoRE improves performance by 3.44\% across all models.
Specifically, traditional methods like DIN-SQL and DAIL-SQL gain 3.5\% and 3.6\% in accuracy, respectively.
Note that DAIL-SQL uses few-shot examples extracted from the training set without retrieval, while MAC-SQL and DIN-SQL employ fixed examples in the prompt template without additional positive or negative signals to indicate relevance. 
Moreover, CoRE provides substantial improvements for advanced models. It enhances MAC-SQL and its GPT-4 version by 5.9\% and 5.0\%, respectively, as the more powerful language model effectively captures contrastive contexts.

The example selection in DAIL-SQL masks domain-specific words in both target and candidate questions, ranking candidates by cosine similarity of their embeddings to pick the most relevant examples for ICL. In contrast, \textbf{our CoRE retrieves a wider range of data, providing diversity and broader coverage} (Figure~\ref{fig:fig3_reward_distribution}) — \textit{answer to RQ-(1)}.

\begin{figure}[h]
  \hspace{0.1cm}
  \includegraphics[width=\columnwidth]{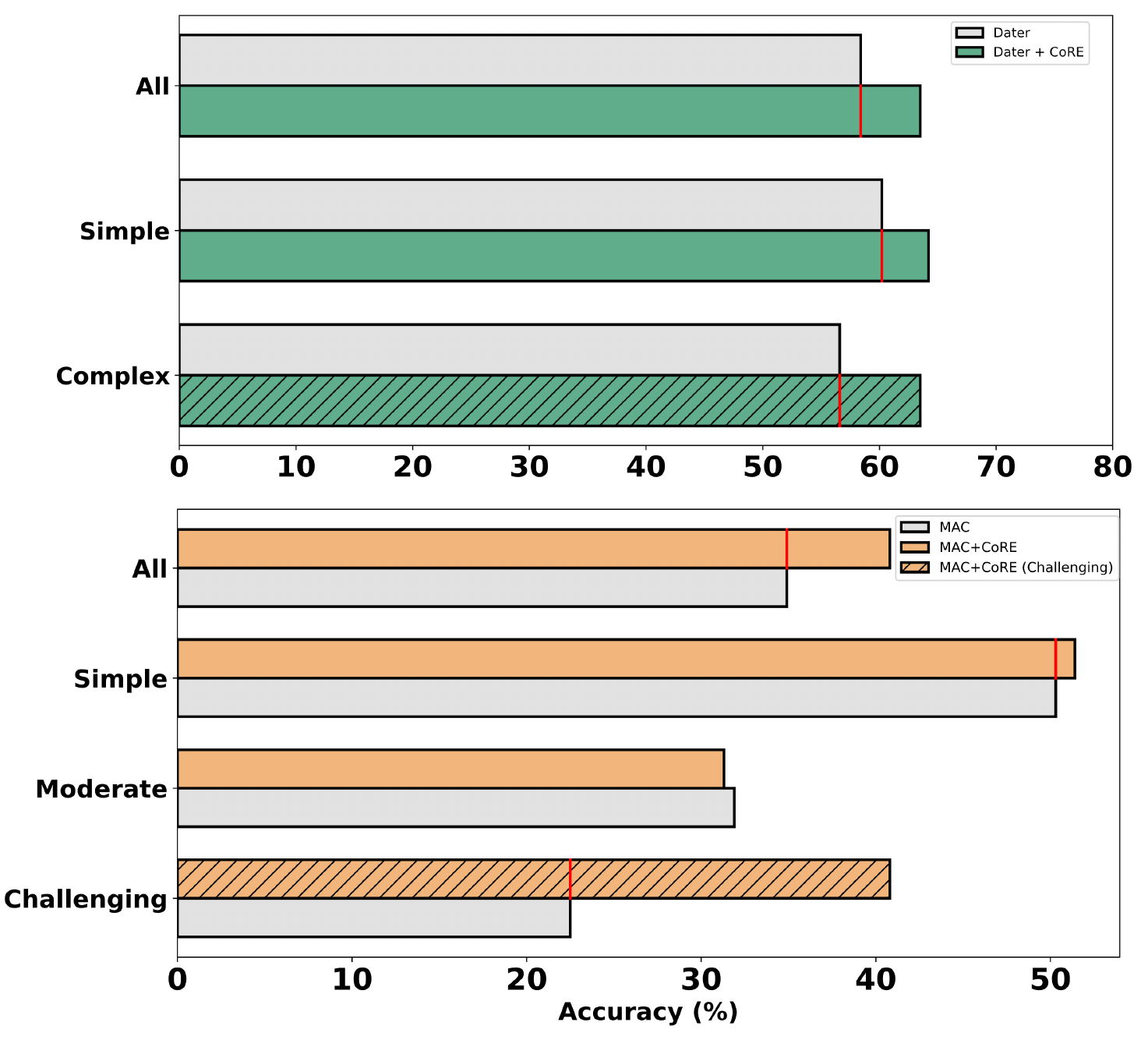}
  \caption{Performance under different difficulty.}
  \label{fig:fig4_difficulty_bird}
  \vspace{-5pt}
\end{figure}

However, it is noteworthy that CoRE slightly reduces the accuracy for DAIL-SQL (SC) by 0.2\%. This result may be due to the Contrastive Thinker in CoRE, which includes a large amount of contextual learning information. In contrast, Self-Consistency~(SC) uses voting on five answers generated with a temperature of 1.0. This high-temperature setting can amplify the impact of noise from excessive contextual information in the generated content. As a result, it leads to negative outcomes. This also suggests that \textbf{combining different prompting methods is not always better}, as CoRE can be viewed as a dynamic few-shot approach for prompt engineering. Moreover, CoRE shows significant improvement for challenging Text-to-SQL tasks as shown in Figure~\ref{fig:fig4_difficulty_bird}. It nearly doubles the execution accuracy, reaching 40\%, compared to MAC-SQL. This also indicates that \textbf{CoRE is useful for challenging knowledge-intensive reasoning questions.}

{
\footnotesize
\begin{table}[!h]
    \vspace{-5pt}
    \caption{Results by GPT-3.5, evaluated by Accuracy on WikiTQ and FinQA Dev.}
    \label{tab:wikitq_results}
    \centering
    \small
    \begin{tabular}{l|l|c}
      \toprule
      \textbf{Dataset} & \textbf{Method} & \textbf{Accuracy~($\%$)} \\
      \midrule
      \multirow{7}{*}{WikiTQ} & StructGPT~\cite{structgpt}       &  64.4 \\
       & \hspace{1cm}+ our CoRE       &  66.1~($\uparrow 1.7$) \\

       & Dater~\cite{dater}    &  58.4 \\
       & \hspace{1cm}+ our CoRE       &  63.5~($\uparrow 5.1$) \\
      \midrule
      \multicolumn{2}{l|}{\textbf{$\Delta$ CoRE}}  & \textbf{3.4} \\
      \midrule
      \multirow{7}{*}{FinQA} & StructGPT~\cite{structgpt}       &  51.2 \\
       & \hspace{1cm}+ our CoRE       &  53.1~($\uparrow 1.9$) \\

       & Dater~\cite{dater}    &  52.4 \\
       & \hspace{1cm}+ our CoRE       &  59.0~($\uparrow 6.6$) \\
      \midrule
      \multicolumn{2}{l|}{\textbf{$\Delta$ CoRE}}  & \textbf{4.25} \\
      \bottomrule
    \end{tabular}
    \vspace{-10pt}
\end{table}
}

\begin{table*}[t!]
\vspace{-10pt}
\caption{Ablation Experiment by GPT-4 (2-shot results.)}
\label{tab:ablation_study_2_shot}
\centering
\setlength{\tabcolsep}{2mm}  
\small
\begin{tabular}{@{}cccccc@{}}
\toprule
\multirow{2}{*}{\textbf{Method}} & \multicolumn{5}{c}{\textbf{2-shot}} \\
\cmidrule(lr){2-6}
& \textbf{\normal \hspace{0.015cm} \normal} & \textbf{\posi \hspace{0.015cm} \posi} & \textbf{\nega \hspace{0.015cm} \nega} & \textbf{\posi \hspace{0.015cm} \nega} & \textbf{\posi$\mid$\nega} \\
\midrule
\multicolumn{1}{@{}c}{\textit{Generic Reasoning}} \\
\quad End-to-End QA  + our CoRE & 67.68 & 67.89~($\uparrow 0.21$) & 67.33~($\downarrow 0.35$) & 68.20~($\uparrow 0.52$) & 68.14~($\uparrow 0.46$) \\
\quad CoT QA~\citep{wei2022chain}  + our CoRE & 68.09 & 68.07~($\downarrow 0.02$) & 67.73~($\downarrow 0.36$) & 68.35~($\uparrow 0.26$) & 68.61~($\uparrow 0.52$) \\

\cmidrule(lr){1-6}
$\Delta$ CoRE & -- & -- & +0.39 & +0.49 \\
\midrule

\multicolumn{1}{@{}c}{\textit{Structure Knowledge Reasoning}} \\
\quad MAC-SQL~\citep{mac-sql} + our CoRE & 56.40 & 55.74~($\downarrow 0.66$) & 45.20~($\downarrow 11.20$) & \textbf{58.92~($\uparrow 2.52$)} & 58.08~($\uparrow 1.68$) \\

\quad Dater~\citep{dater} + our CoRE & 64.88 & 66.20~($\uparrow 1.32$) & 65.27~($\uparrow 0.39$) & 68.03~($\uparrow 3.15$) & \textbf{69.72~($\uparrow 4.84$)}  \\

\cmidrule(lr){1-6}
$\Delta$ CoRE & -- & -- & -- & +2.835 & +3.260 \\
\bottomrule
\end{tabular}
\vspace{-8pt}
\end{table*}

\begin{table}[t]
\caption{Ablation Study with GPT-4 (0/1-shot Settings)}
\label{tab:ablation_study_0_1_shot}
\centering
\small
\setlength{\tabcolsep}{0.5mm}
\begin{tabular}{@{}p{2cm}cccc@{}}
\toprule
\multirow{2}{*}{\textbf{Method}} & \multicolumn{1}{c}{\textbf{0-shot}} & \multicolumn{3}{c}{\textbf{1-shot}} \\
\cmidrule(lr){2-2} \cmidrule(lr){3-5}
& \textbf{\normal} & \textbf{\normal} & \textbf{\posi} & \textbf{\nega} \\
\midrule
End-to-End QA\\\raggedleft + our CoRE & 66.79 & 66.83 & 66.60~($\downarrow$0.23) & 66.04~($\downarrow$0.79) \\
CoT QA\\\raggedleft + our CoRE & 67.68 & 67.85 & 68.71~($\uparrow$0.86) & 66.69~($\downarrow$1.16) \\
MAC-SQL\\\raggedleft + our CoRE & 48.77 & 54.36 & 54.45~($\uparrow$0.09) & 40.10~($\downarrow$14.26) \\
Dater\\\raggedleft + our CoRE & 60.19 & 64.90 & 66.38~($\uparrow$1.48) & 64.82~($\downarrow$0.08) \\
\bottomrule
\end{tabular}
\vspace{-10pt}
\end{table}

\paragraph{TableQA}

Consistent with the Text-to-SQ, we employed the 2-shot setting. Table~\ref{tab:wikitq_results} shows that when applied to TableQA baselines, CoRE boosts StructGPT’s accuracy by 1.7\% and improves Dater’s accuracy from 58.4\% to 63.5\% ($\uparrow$ 5.1\%). Both StructGPT and Dater use fixed few-shot examples in their \textit{Iterative Reading-then-Reasoning} and \textit{Jointly Reasoning} methods. It can be observed that using the CoRE plugin on Dater results in greater improvement. \textbf{This difference stems from the grounding phase in reasoning}. 
Dater leverages its Evidence Decomposer to break down and filter high-density structured knowledge, producing a cleaner reasoning input $\mathcal{E}$ with reduced noise. Consequently, with this refined $\mathcal{E}$, CoRE performs Contrastive In-Context Learning more effectively, resulting in a larger boost in reasoning performance. Overall, CoRE yields an average improvement of 3.4\% on the WikiTQ dataset, and as shown in Figure~\ref{fig:fig4_difficulty_bird}, gains on complex issues dominate the overall enhancement.

On FinQA, CoRE increases StructGPT’s accuracy by 1.9\% and Dater’s by 6.6\%, achieving an average gain of 4.25\%. As FinQA contains much less structured knowledge than WikiTQ, grounding and reasoning play a stronger role. Moreover, the higher plain-text content in FinQA simplifies ICL, allowing CoRE to enhance relevant information more effectively and attain a higher success rate.

\subsection{Ablation Study}\label{Exp:ablation_study}
We analyzed the differences between existing methods and our CoRE, then proposed the our research questions (\S\ref{sec:preliminary}). The \textit{Answer to RQ-(1)}: our MCTS-generated Experience Memory Dataset offers greater diversity and coverage than datasets relying solely on training data by comparing with DAIL-SQL in \S~\ref{Exp:result_and_analysis}.
To answer the remaining RQs, we further conducted ablation studies on MAC-SQL and Dater using GPT-4, demonstrating that CoRE benefits both structured and generic reasoning. To confirm its effectiveness, we compared the fixed few-shot method used in baselines~\cite{mac-sql,dater} with our Retriever-provided positive and negative examples across 0-shot to 2-shot settings. As shown in Table~\ref{tab:ablation_study_0_1_shot} and Table~\ref{tab:ablation_study_2_shot}, \raisebox{-0.2ex}{\normal} denotes fixed examples without any additional signals, while \raisebox{-0.2ex}{\posi} and \raisebox{-0.2ex}{\nega} represent positive and negative cases for the new question. These three types of examples are provided as N-shot inputs for reasoning in a single round. However, "\raisebox{-0.2ex}{\posi} $\mid$ \raisebox{-0.2ex}{\nega}" uses a two-round approach, with positive cases guiding the first and negatives correcting errors in the second. The results lead to the following conclusions:

\vspace{5pt}
\noindent \textbf{RAG helps effective knowledge augmentation}. — \textit{Answer to RQ-(2)}.
For structure knowledge~(high-density) reasoning, RAG proves to be an effective approach for diluting intensive knowledge with effective examples, to guide LLM to interpret complex knowledge. This is consistent with the effectiveness demonstrated by DAIL-SQL. Compared with the fixed example without specifically associated signals, CoRE can improve performance through dynamically recalled positive examples (without comparison), which can be seen in both 1-shot ($\uparrow 1.48\%$) and 2-shot ($\uparrow 1.32 \%$). 
The reasoning task in the more general FinQA leans towards a strong LLM background, showing similar performance without much corresponding increase.

%
    

\noindent \textbf{Contrastive ICL brings consistent improvement} — \textit{Answer to RQ-(3)}.
Leveraging CoRE's dynamic retrieval to obtain relevant examples, contrastive ICL outperforms both positive-positive (\raisebox{-0.2ex}{\posi} \raisebox{-0.2ex}{\posi}) and negative-negative (\raisebox{-0.2ex}{\nega} \hspace{0.001cm} \raisebox{-0.2ex}{\nega}) combinations. In particular, positive-negative combinations (\raisebox{-0.2ex}{\posi} \raisebox{-0.2ex}{\nega} or \raisebox{-0.2ex}{\posi} $\mid$ \raisebox{-0.2ex}{\nega}) yield the highest gains—up to $\uparrow 2.835\%$ in a single round or $\uparrow 3.260\%$ across two rounds. This consistent improvement in both generic and structured knowledge reasoning demonstrates that contrastive ICL effectively stabilizes and enhances performance on challenging tasks.


\paragraph{Only using negative examples without pointing out the mistakes would harm performance}
The data indicates that using positive examples before generation yields better outcomes, while negative examples can have adverse effects. In the \textbf{2-shot} columns of the table, positive-positive combinations (\raisebox{-0.2ex}{\posi}  \raisebox{-0.2ex}{\posi}) show superior performance compared to fix examples~(\raisebox{-0.2ex}{\normal}  \raisebox{-0.2ex}{\normal}). Conversely, negative-negative combinations (\raisebox{-0.2ex}{\nega} \raisebox{-0.2ex}{\nega}) degrade model performance, showing that negative examples introduce noise and negatively affect results. It seems that only using negative examples without pointing out the mistakes would harm performance, as generative models will involuntarily imitate incorrect cases even if they have been informed that these are wrong cases.

\paragraph{Different ways of comparison in the dialogue yield similar effects.}
In the results of using Contrastive In-Context Learning, the effect of comparing in single-turn dialogues (\raisebox{-0.2ex}{\posi} \raisebox{-0.2ex}{\nega}) and comparing before and after in two-turn dialogues (\raisebox{-0.2ex}{\posi} $\mid $ \raisebox{-0.2ex}{\nega}) is competitive. This means that our CoRE can be implemented in a more flexible way, and even with limited context length, similar improvements can still be achieved through multi-turn dialogues.

We experimented with various strategies in prompt engineering for our CoRE. In the Text-to-SQL task, we integrated CoRE into the Self-Consistency~(SC) baseline method but observed negative effects (\S~\ref{result:text2sql}), leading us to discontinue discussing this strategy. In the Text-to-SQL task, we used LlaMA-3 to apply CoRE on MAC-SQL multiple times for iterative optimization in multi-turn dialogues, a process known as bootstrapping.

\subsection{More Discussion}\label{Exp:strategy_and_discussion}
\begin{figure}[h!]
  \vspace{-10pt}
  \includegraphics[width=\columnwidth]{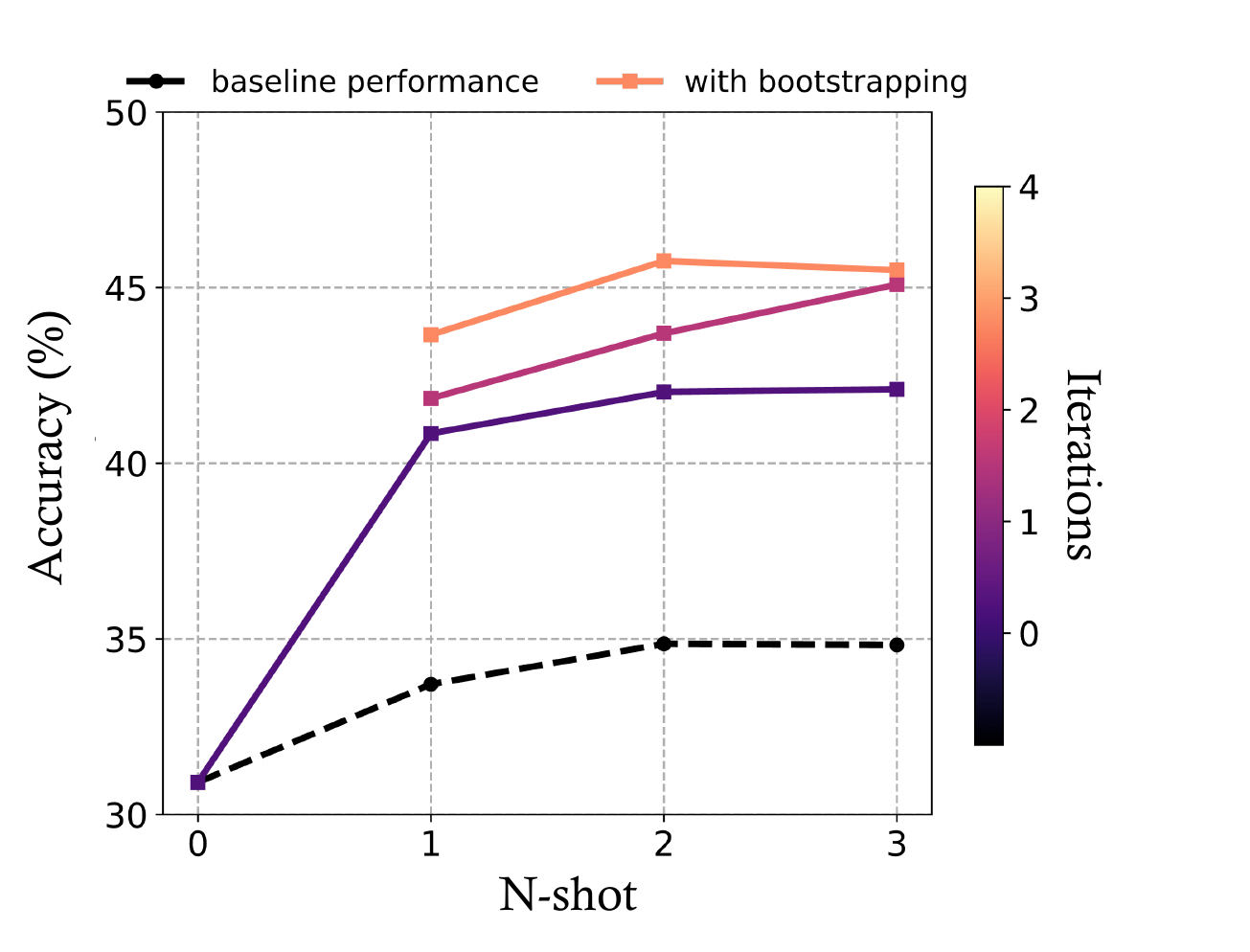}
  \caption{\small Bootstrapping with CoRE by LlaMA-3-70b.}
  \label{fig:fig5_bootstrap_analysis}
  \vspace{-10pt}
\end{figure}
\vspace{2pt}
\paragraph{Bootstrapping.}\label{strategy:bootstrapping}
We repeatedly apply CoRE in multi-turn dialogues under various few-shot settings—one iteration equals one use of CoRE and N-shot indicates the number of contrastive examples in context. Due to context length limits, structured knowledge $\mathcal{E}$ is included only in the first round, with subsequent iterations updating the current answer solely based on retrieved examples. As shown in Figure~\ref{fig:fig5_bootstrap_analysis}, while the baseline improves modestly from 30.92\% (0-shot) to 34.83\% (3-shot), bootstrapping achieves 42.11\% at 3-shot without iterations, rising further to 45.09\% at 3-shot after one iteration and 45.76\% at 2-shot after two iterations. This shows that using bootstrapping can lead to more accurate results when there is a sufficient budget and low requirements for delay restrictions.



\begin{figure}[h]
  \vspace{-10pt}
  \includegraphics[width=\columnwidth]{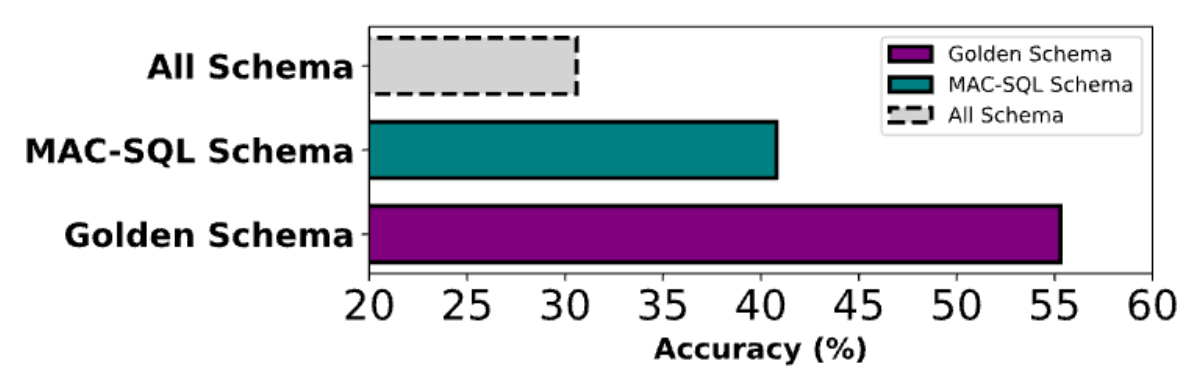}
  \caption{Different Schema for CoRE by LlaMA-3-70b}
  \label{fig:fig6_grounding}
  \vspace{-10pt}
\end{figure}

\vspace{2pt}
\paragraph{Grounding and Reasoning.} 
In our work, we concentrate on the reasoning aspect of structured knowledge question-answering by following the grounding method of each baseline. We wonder how the grounding results affect our CoRE reasoning, so we also experimented with using a golden schema for CoRE with MAC-SQL. 
The results in Figure~\ref{fig:fig6_grounding} show that the grounding process will significantly affect the results of CoRE reasoning. This observation is consistent with the phenomenon observed on two different baselines in TableQA. Using CoRE as a dynamic few-shot strategy on more fine-grained structured knowledge $\mathcal{E}$ can lead to higher returns.

\paragraph{Computational Analysis} 
CoRE’s one-time offline cost is amortized across tasks, its inference-time overhead is minimal, and it remains practical under strict token constraints. We analyze three aspects here.
\textit{\textbf{1. Offline Investment:} } 
We run Monte Carlo Tree Search once per training example to build the Experience Memory of (question, context, answer) triplets. This step produces a reusable repository without further model training and resolves cold-start issues. After construction, no additional offline computation is required for any downstream task.
\textit{\textbf{2. Inference-Time Retrieval Overhead:}}
At inference, each query is mapped to an embedding and compared against stored entries. Retrieving the top-$k$ candidates requires computing similarity scores for all entries (complexity $O(N\times d)$) and re-ranking the best matches (complexity $O(N\log k)$). In practice, retrieval plus prompt construction takes under 60 ms per query, which is negligible compared to typical model generation time.
\textit{\textbf{3. Scalability under Token Limits (Few-Shot Adaptability):}}
CoRE supports multi-turn in-context learning (like \raisebox{-0.2ex}{\posi} $\mid $ \raisebox{-0.2ex}{\nega}), allowing us to present contrastive examples across several shorter prompts rather than one long prompt. This design reduces the impact of strict token limits while preserving few-shot performance. Bootstrapping further refines examples over turns, so even a small set of positive and negative instances yields consistent improvements without excessively long context.

\vspace{-5pt}
\section{Related Work}

\vspace{-3pt}
\subsection{Contrastive learning}
Contrastive learning~\cite{contrasive_learning} is a well-known technique that encourages models to differentiate between "positive" and "negative" samples, commonly used in areas like RLHF, where reward models are trained on human preference data~\cite{prm800k,rlhf}. However, contrastive methods are less explored in prompt engineering. It remains unclear whether contrastive demonstrations can enhance the reasoning process~\cite{understanding_cot_prompting}, though recent work shows that contrastive CoT consistently improves In-context learning~\cite{contrastive_cot}.

\subsection{Retrieval Augmented Generation}

Retrieval Augmented Generation (RAG) enhances LLMs by fetching relevant documents from external databases and integrating them into the generation process~\citep{gao2023retrieval, lewis2020retrieval, khandelwal2020generalization, izacard2021leveraging, luo2023sail} (see Figure~\ref{fig:fig1_rag_vs_core}(a)). Recent work has focused on helping LLMs decide when and what to retrieve~\citep{schick2024toolformer} or on better exploiting context~\citep{sarthi2024raptor,kim2024sure}. While Self-Reflective RAG~\citep{asai2023self} uses reflection tokens to guide retrieval and Corrective RAG~\citep{yan2024corrective} proposes a lightweight evaluator (albeit lacking high-level reasoning), CoRE instead emphasizes effective utilization of retrieved content and contexts by contrastive learning.

\subsection{Structured Knowledge Reasoning}
Structured knowledge, such as tables, knowledge graphs, and databases, has been central to knowledge grounding~\cite{liu-etal-2025-solid} and reasoning tasks. Due to heterogeneous data formats, methods~\cite{xie-etal-2024-decomposition} have emerged that leverage specific training setups to learn these representations. However, recent prompt-based methods, capitalizing on LLMs' capabilities, aim to process structured knowledge without re-training. Approaches like MAC-SQL~\cite{mac-sql}, StructGPT~\cite{structgpt}, Dater~\cite{dater}, DIN-SQL~\cite{din-sql}, and DAIL-SQL~\cite{dail-sql} enhance LLMs' reasoning through multi-agent frameworks, decomposition, and schema linking, improving performance with techniques like evidence decomposition and self-correction.

\vspace{-5pt}
\section{Conclusion}
We have introduced CoRE, a training-free contrastive RAG framework that uses Monte Carlo Tree Search to build an Experience Memory for structured knowledge reasoning. Our experiments on Text-to-SQL and TableQA tasks demonstrate that CoRE improves performance substantially on challenging queries by retrieving evidence with greater diversity and coverage than prior approaches. By incorporating both positive and negative examples through contrastive in-context learning, CoRE enhances reasoning robustness across different grounding methods and remains scalable under strict token limits through multi-turn prompting. The offline memory construction incurs a one-time cost that is amortized across tasks, and inference-time retrieval adds negligible overhead. These results establish CoRE as a flexible, efficient solution that extends the generalization of NLP models to complex structured reasoning scenarios.

\section*{Limitation}
While CoRE provides a promising solution for the cold start problem in RAG tasks and offers a tailored approach for structured knowledge reasoning, its performance may vary depending on the coverage and quality of the Experience Memory dataset. The effectiveness of the method relies heavily on the breadth of the data used to generate this memory, and when transferring the approach to a new task, the Experience Memory must be rebuilt as per the outlined method for stable results.
Additionally, CoRE’s performance is highly dependent on the accuracy of the initial grounding phase. The grounding phase identifies relevant data from structured sources, and inaccuracies in this process may hinder the model's ability to reason effectively. This aspect requires careful attention, especially when applied to real-world scenarios where structured knowledge might be incomplete or ambiguous.

\bibliography{custom}

\appendix
\onecolumn

\section{More Related Work}

\subsection{Mental Representations \& Memory Representations}\label{MR}
This concept of\textit{ Mental Representations}~\cite{peak_deliberate_learning} on human cognition refers to a mental structure that corresponds to an object, an idea, or a collection of information, whether concrete or abstract, that the brain is contemplating. \textit{Mental Representations} plays a crucial role across various domains, facilitating effective planning and problem-solving, which can be figuratively stated as:
\begin{displayquote}
\textit{While the \textit{Mental Representations} give masters a view of the forest when everyone else sees only trees, they also allow masters to zero in on the trees when necessary.}
\begin{flushright}
------\textit{\citeauthor{peak_deliberate_learning}}
\end{flushright}
\end{displayquote}
Moreover, from the perspective of natural language development, the meaning of language is not right over there to see, to grasp or to learn, but rather constructed during a communicative event through interactive processes among interlocutors. Hence, intuitively, developing structures similar to \textit{Mental Representations} within LLMs' dialogue contexts naturally corresponds to the internal cognitive mechanisms of language, while also capturing the essential features of \textit{Mental Representations}, including its dynamic formation and continual refinement.

Thus, intuitively, constructing structures akin to Mental Representations within the context of dialogue, significantly enhances the understanding and reasoning of natural language content. 
Consequently, we borrow the concept of \textit{Mental Representations} and propose \textit{\textbf{Memory Representations}} for LLMs, which enhances their abilities in comprehending, memorizing, organizing, analyzing, and ultimately making decisions with interactive information. Many paradigms in reasoning and planning incorporate components with functionalities similar to \textit{\textbf{Memory Representations}}, albeit in naive forms, such as long-short-term memory~\cite{shinn2023reflexion}, task-specific heuristic function and policy networks~\cite{XoT}. However, they are more than just storage memory in different forms or functions; indeed, they should encapsulate valid experiences that can guide efficient practices in subsequent instances. They optimizes decision-making policies through policy gradient methods relying on binary or scalar feedback, which significantly differs from the human-like decision-making process. However, quantifying the reward from natural language text is particularly challenging due to its inherent ambiguity and the subjective nature of language interpretation. Additionally, many existing methods~\cite{shinn2023reflexion, XoT} equip LLMs with crucial human capabilities like memory, thereby regarding them as autonomous agents to complete human-like reasoning and planning. In this paper, we concentrate on the task of decision-making through natural language text, which encapsulates a comprehensive range of information, nuances, and potential actions.

\subsection{Experiential Learning}
To improve the performance of LLMs, researchers provide textual experience to LLMs through prompts. Early studies primarily involve manually crafting such experiential prompts~\citep{wei2022chain,kong2023better}, while more recent work focuses on utilizing the LLMs themselves to obtain task-solving experience automatically. \citet{zhao2023expel} and \citet{chen2024grimoire} leveraged LLMs to automatically summarize experience from manually annotated NLP datasets.
\citet{zhao2023expel} employed Reflexion \citep{shinn2023reflexion} to generate reasoning chains for each question. Then, the experience is summarized from the questions, chains, and human-annotated labels by LLMs.
\citet{chen2024grimoire} analyzed the impact of different examples and prompts on the quality of the summarized experience.
However, these methods still require human labor to obtain experience and determine which experience to employ for the current question. Our framework autonomously learns and selects experience by RAG, which is continual and more consistency.

\section{Experiment Details}\label{app:exp_details}

\subsection{Pseudo-code}
\begin{algorithm}[h]
\centering
\caption{MCTS with LLMs}\label{algo:tree-based_search}
\begin{minipage}{\linewidth} 
\small
\begin{algorithmic}[1]
    \Require Overall question $\mathcal{Q}$, structured evidence $\mathcal{E}$, large language model Generator, action generator $\mathcal{G}(\cdot)$, number of generated sub-questions $d$, depth limit $L$, number of roll-outs $N$, and exploration weight $w$, reward function to return specific reward $f_{r}(s_{k-1}, q_k)$
    \For {$n \gets 0, \dots, N - 1$}
        \State $k \gets 1$
            
        \While {$N(s_{k-1}) > 0$} \Comment{Selection}
            \State $N(s_{k-1}) \gets N(s_{k-1}) + 1$
            \State $q_k \gets \arg\max_{q \in \text{sub}(q_{k-1})}  \left[ Q_\text{value}(s_{k-1}, q) + w \sqrt{\frac{\ln N(s_{k-1})}{N(c(s_{k-1}, q))}} \right]$
            \State $r_k = f_{r}(s_{k-1}, q_k)$, $s_{k} \gets c(s_{k-1}, q_k)$
            \State $k \gets k + 1$
        \EndWhile
        \While {$s_k$ is not a terminal state $\wedge$ $k \leq L$}
            \For {$i \gets 1, \dots, d$} \Comment{Expansion}
                \State  sample $\text{sub}(q_{k-1}) \gets \text{sub}(q_{k-1}) \cup \{\mathcal{G}(q|s_{k-1}^*, \dots, s_0, \mathcal{E})\}$, \\ \Comment{\textit{Invoking LLM}}
            \EndFor
            \State $q_{k}^* \gets \arg \max_{q \in \text{sub}(q_{k-1})} f_{r}(s_{k-1}, q_k)$ \Comment{Simulation}
            \State $r_k \gets r(s_{k-1}, q_k)$, $s_{k} \gets c(s_{k-1}, q_k)$
            \State $k \gets k + 1$
        \EndWhile
        \For {$k' \gets (k-1), \dots, 0$} \Comment{Back propagation}
            \State Update $Q_{\text{value}}(s_{(k-1)'}, a_{k'})$ with $\{r_{k'}, r_{k'+1}, \dots, r_{k-1}\}$
        \EndFor
    \EndFor
\end{algorithmic}
\end{minipage}
\end{algorithm}

\subsection{Experience Memory Dataset Details}
Finally, we build the Experience Memory Dataset on training part~(Table~\ref{tab:dataset_stats}) and finally obtain 85,956 cases with Bird~\cite{bird}, 98,586 cases with WikiTQ~\cite{wiki_table_question} and 55,399 cases with FinQA~\cite{finqa}, which has expanded the original training dataset by 8-9 $\times$.

\begin{figure}[h]
  \centering
  \includegraphics[width=0.7\columnwidth]{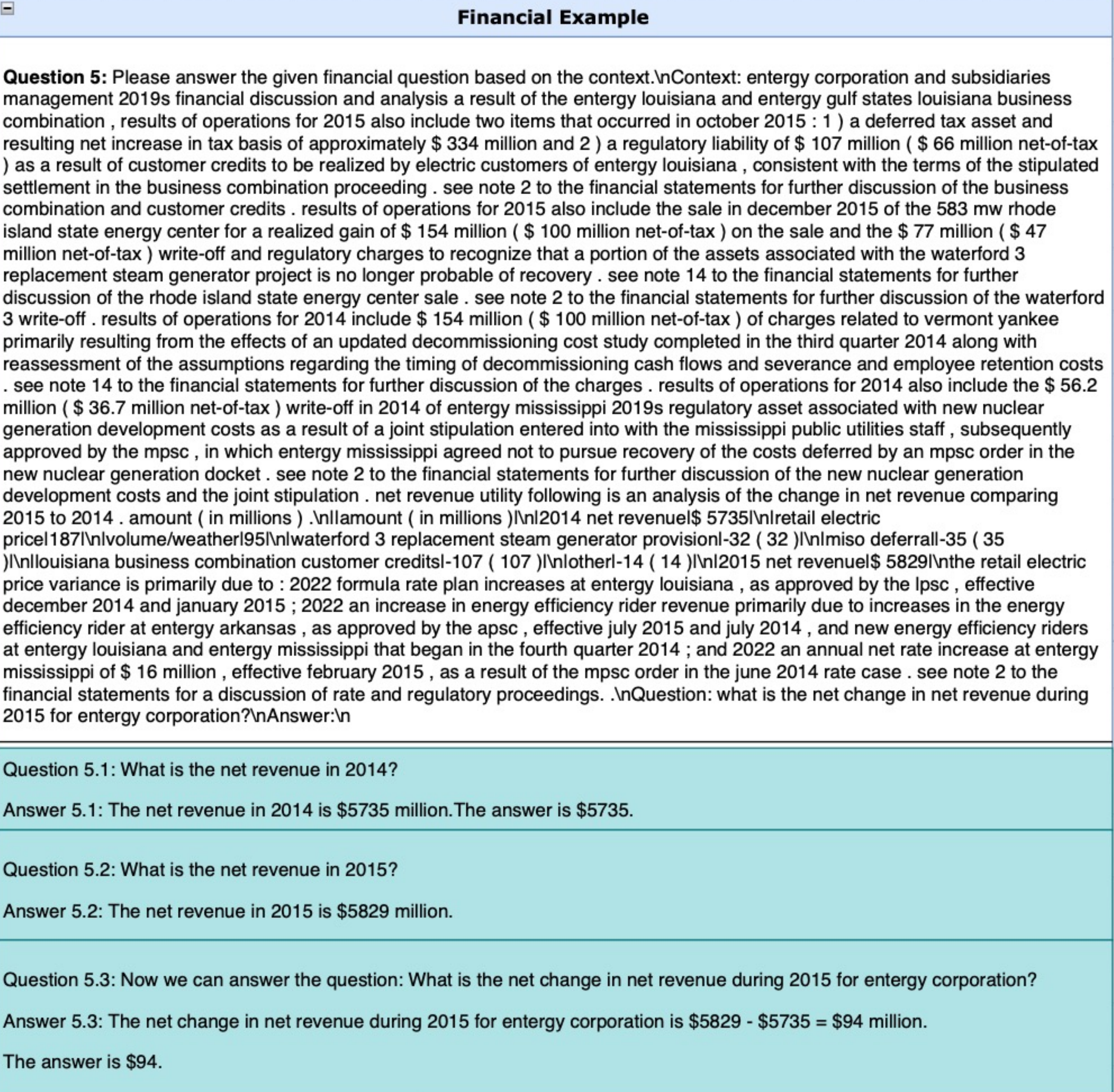}
  \caption{A trajectory of Experience Memory Builder for FinQA.}
  \label{fig:finqa_mcts_example}
\end{figure}

As we stated in \S~\ref{method:experience_memory}, structured knowledge reasoning can be solved by planning for decomposition. Figure~\ref{fig:finqa_mcts_example} shows one trajectory for FinQA Dataset. 

\subsection{Datasets}\label{apd:datasets}

\begin{table}[h]

\centering
\begin{tabular}{c|cc|ccc|ccc}
\toprule
&\multicolumn{2}{c|}{Overall Length}&\multicolumn{3}{c|}{Train}&  \multicolumn{3}{c}{Dev/Test}\\
\midrule
Dataset &
  \makecell[l]{In. \\ (avg)} &
  \makecell[l]{Out. \\ (avg)} &
  Cnt &
  \makecell[l]{In. \\ (max)} &
  \makecell[l]{Out. \\ (max)} &
  Cnt &
  \makecell[l]{In. \\ (max)} &
  \makecell[l]{Out. \\ (max)} \\
\midrule

\defcitealias{pasupat-liang-2015-compositional}{WikiTQ}\citetalias{pasupat-liang-2015-compositional} & 831.8 & 5.8 & 11321 & 2028 & 273 & 4344 & 2048 & 148 \\

\defcitealias{li2023llm}{BIRD}\citetalias{li2023llm} & 439.8 & 63.3 & 9428 & 1992 & 347 & 1534 & 1214 & 386 \\

\defcitealias{chen2021finqa}{FinQA}\citetalias{chen2021finqa} & 1230.3 & 21.0 & 6251 & 2040 & 72 & 1147 & 2048 & 61 \\
\bottomrule
\end{tabular}
\caption{Token sequence length statistics for each dataset.}
\label{tab:dataset_stats}
\end{table}

\paragraph{TableQA}
We evaluate our proposed CoRE on two table-based reasoning datasets, including  WikiTableQuestion \citep{wiki_table_question}, and FinQA \citep{finqa}. The details of these two datasets are provided as follows.

\begin{itemize}
    \item \textbf{WikiTableQuestion} contains complex questions annotated by crowd workers based on Wikipedia tables. The crowd workers are asked to write questions that involve several complex operations such as comparison, aggregation and arithmetic operations, which require compositional reasoning over a series of entries in the given table. We use the standard test set with 4,344 samples.
    \item \textbf{FinQA} consists of complex questions curated by financial experts based on real-world financial reports from S\&P 500 companies. These questions involve intricate numerical reasoning, requiring multi-step operations such as addition, subtraction, and ratio calculations across both structured tables and unstructured text.  The dataset is split into a training set with 6,251 samples, a validation set with 883 samples, and a test set with 1,147 samples.

\end{itemize}

\paragraph{Text-to-SQL}

\begin{itemize}
    \item \textbf{Bird} dataset is extensive, with 12,751 unique question-SQL pairs from 95 large-scale databases totaling 33.4 GB. It encompasses a wide array of over 37 professional domains including blockchain, hockey, healthcare, and education. To assist models in generating accurate SQL queries, Bird incorporates external knowledge from four sources: numerical reasoning, domain-specific information, synonyms, and value examples.
\end{itemize}

\subsection{Baselines}\label{apd:baselines}

To ensure a stable generation, we set the temperature to 0 when reasoning with CoRE. The baselines we used are as follow:

\begin{itemize}
\item {\textbf{MAC-SQL }} MAC-SQL~\cite{mac-sql} introduces a novel multi-agent framework for the Text-to-SQL task, leveraging large language models (LLMs). The framework comprises three specialized agents: the Selector, responsible for condensing the database by retaining only relevant table schemas; the Decomposer, which breaks down complex questions into simpler sub-problems; and the Refiner, which validates and refines the generated SQL queries. The Decomposer first determines whether the question can be addressed using a simple SQL query. When the question requires a more complex approach, it applies a chain-of-thought (CoT) strategy to generate sub-questions and corresponding SQL queries, iterating until the final SQL query is constructed. Throughout this process—whether generating SQL directly or decomposing the question—few-shot learning techniques are used to ensure precise results. 

\item {\textbf{StructGPT}} StructGPT~\cite{structgpt} employs an iterative reading-then-reasoning (IRR) framework to enhance the capability of large language models (LLMs) in reasoning over structured data, particularly tables. The method involves constructing specialized interfaces to access and filter relevant evidence from the table data efficiently. Through an iterative process of invoking these interfaces to extract information, linearizing the extracted data into a readable format for LLMs, and generating responses based on this linearized data, the framework progressively homes in on the target answers to given queries. This approach allows LLMs to leverage their reasoning abilities effectively by focusing on the reasoning task rather than the intricacies of understanding structured data formats.

\item {\textbf{Dater}} Dater~\cite{dater} leverages large language models (LLMs) to decompose table-based reasoning tasks into more manageable sub-tasks. DATER addresses two main challenges in table-based reasoning: dealing with large evidence, such as extensive tables with numerous rows and columns, and reasoning over complex questions that require multi-step logical deductions. The method involves an Evidence Decomposer to break down the vast table into relevant sub-evidence by predicting the indexes of rows and columns, and a Question Decomposer that simplifies complex questions into a series of simpler sub-questions using a "parsing-execution-filling" strategy with intermediate SQL queries. These decomposed elements are then used by the LLM to reason and derive the final answer, enhancing the performance and interpretability of table-based reasoning.

\item {\textbf{DIN-SQL}} DIN-SQL~\cite{din-sql} enhances reasoning capabilities by decomposing the text-to-SQL task into smaller sub-tasks, which are then solved sequentially and their solutions fed into the LLMs. This includes modules for schema linking, query classification and decomposition, SQL generation, and self-correction. DIN-SQL employs prompting strategies tailored to the complexity of the task, addressing challenges in schema linking and utilizing LLMs for self-correction of minor mistakes in generated SQL queries. Through extensive experiments on the Spider and BIRD datasets, DIN-SQL demonstrates significant performance improvements, pushing the accuracy of LLMs towards or even surpassing the state-of-the-art fine-tuned models.

\item {\textbf{DAIL-SQL}} DAIL-SQL~\cite{dail-sql} integrates structured knowledge as SQL statements, selects examples based on skeleton similarities, and removes cross-domain knowledge for token efficiency. DAIL also explores the potential of open-source LLMs through supervised fine-tuning, demonstrating significant performance improvements. In the context of Few-shot learning, the paper introduces DAIL Selection, a strategy that considers both question and query similarities for selecting the most helpful examples, and DAIL Organization, a method that presents both questions and corresponding SQL queries to retain the question-SQL mapping while improving token efficiency. These approaches collectively contribute to the state-of-the-art performance of DAIL-SQL in Text-to-SQL reasoning, especially in scenarios with limited examples.

\end{itemize}

\section{Prompt Details}\label{apd:prompt_Details}
We provide the prompt details in MAC-SQL + CoRE as an example.

\begin{tcolorbox}[colback=lightgray!10, colframe=black, title={Prompt Details}]
\textbf{[Instruction]} \\
When executing SQL below, some errors occurred. Please fix SQL based on query and database info. Solve the task step by step if necessary. Use SQL format in the code block, and indicate script type in the code block. When you find an answer, verify it carefully. Include verifiable evidence in your response if possible.

\textbf{[Constraints]}  \\
- In `SELECT <column>`, just select needed columns in the \textbf{[Question]} without any unnecessary column or value. \\
- In `FROM <table>` or `JOIN <table>`, do not include unnecessary tables. \\
- If using max or min functions, `JOIN <table>` FIRST, THEN use `SELECT MAX(<column>)` or `SELECT MIN(<column>)`. \\
- If [Value examples] of <column> has 'None' or None, use `JOIN <table>` or `WHERE <column> IS NOT NULL`. \\
- If using `ORDER BY <column> ASC|DESC`, add `GROUP BY <column>` before selecting distinct values.

\textbf{[Query]} \\
\{query\} \\
\textbf{[Evidence]} \\
\{evidence\} \\
\textbf{[Database info]} \\
\{desc\_str\} \\
\textbf{[Foreign keys]} \\
\{fk\_str\} \\
\textbf{[Old SQL]} \\
\begin{verbatim}
\{sql\}
\end{verbatim}

\textbf{[SQLite Error]} \\
\{sqlite\_error\} \\
\textbf{[Exception Class]} \\
\{exception\_class\} \\

Now please fix the old SQL and generate new SQL again. \\ 
\textbf{[Correct SQL]} 

\end{tcolorbox}

\subsection{Retriever Result Examples}
\begin{itemize}
    \item \textbf{Score}: 0.6489822864532471
    \item \textbf{Reward}: 0.99
    \item \textbf{SQL Query}: 
    \begin{verbatim}
    SELECT count(*) 
    FROM customer_address 
    INNER JOIN address_status 
    ON customer_address.status_id = address_status.status_id 
    WHERE address_status.address_status = 'Inactive'
    \end{verbatim}
    \item \textbf{Natural Language Explanation}: 
    This SQL query is designed to count the number of customer addresses that have an 'Inactive' status. Here's a breakdown of how it works:

    \begin{itemize}
        \item \textbf{SELECT count(*)}: This part of the query specifies that we want to retrieve the total count of rows that meet the specified conditions. The asterisk (*) is a wildcard that stands for "all rows."
        \item \textbf{FROM customer\_address}: This specifies that we are querying from the 'customer\_address' table.
        \item \textbf{INNER JOIN address\_status ON customer\_address.status\_id = address\_status.status\_id}: This performs an inner join between the 'customer\_address' table and the 'address\_status' table based on the 'status\_id' field. For each row in the 'customer\_address' table, it will find a corresponding row in the 'address\_status' table where the 'status\_id' values match.
        \item \textbf{WHERE address\_status.address\_status = 'Inactive'}: This filters the results to include only rows where the 'address\_status' field in the 'address\_status' table is 'Inactive'.
    \end{itemize}

    In summary, this query counts the number of customer addresses that are marked as 'Inactive' in the address status.
\end{itemize}

\begin{itemize}
    \item \textbf{Score}: 0.6261888146400452
    \item \textbf{Reward}: 0.0
    \item \textbf{SQL Query}: 
    {
    \small
    \begin{verbatim}
    SELECT 
        (COUNT(CASE WHEN e.product = 'Credit card' THEN 1 END) / COUNT(*)) * 1.0 AS percentage
    FROM 
        client c 
    LEFT JOIN 
        events e ON c.client_id = e.client_id 
    WHERE 
        c.sex = 'Male';
    \end{verbatim}
    }
    \item \textbf{Natural Language Explanation}: 
    This SQL query is designed to calculate the percentage of events related to 'Credit card' for male clients. Here's a breakdown of how it works:
    
    \begin{itemize}
        \item \textbf{SELECT (COUNT(CASE WHEN e.product = 'Credit card' THEN 1 END) / COUNT(*)) * 1.0 AS percentage}: This part of the query calculates the percentage of events where the product is 'Credit card'. The CASE statement counts the number of rows where the product is 'Credit card', and this count is divided by the total number of rows (COUNT(*)) to get the proportion. Multiplying by 1.0 ensures that the result is a decimal number. The result is given the alias 'percentage'.
        \item \textbf{FROM client c}: This specifies that we are querying from the client table (which has been aliased as c).
        \item \textbf{LEFT JOIN events e ON c.client\_id = e.client\_id}: This joins the client table with the events table based on the client\_id field. It ensures that all rows from the client table are included in the result, even if there is no matching row in the events table.
        \item \textbf{WHERE c.sex = 'Male'}: This filters the results to only include rows where the sex field in the client table is 'Male'.
    \end{itemize}
    In summary, this query calculates the percentage of events related to 'Credit card' for male clients.
\end{itemize}

\vspace{10em}
\subsection{Contrastive Prompt}
\small
\begin{dialogbox}
\begingroup
Given a [Database schema] description, a knowledge [Evidence] and the [Question], you need to use your  <<<<<EXPERIENCE MEMORY>>>>> to deal with the <<<<<NEW CASE>>>>> with correct and executable SQL. Please refer to <Successful Experience> and avoid making the same mistakes as the <Failed Experience>.

When generating SQL, we should always consider constraints:
\textbf{[Constraints]}
\begin{itemize}
    \item In `SELECT <column>`, just select needed columns in the [Question] without any unnecessary column or value.
    \item In `FROM <table>` or `JOIN <table>`, do not include unnecessary table.
    \item If using max or min functions, `JOIN <table>` FIRST, THEN use `SELECT MAX(<column>)` or `SELECT MIN(<column>)`.
    \item If [Value examples] of <column> has 'None' or None, use `JOIN <table>` or `WHERE <column> IS NOT NULL` is better.
    \item If using `ORDER BY <column> ASC|DESC`, add `GROUP BY <column>` before to select distinct values.
\end{itemize}

\textbf{<<<<<EXPERIENCE MEMORY>>>>>}

\textbf{<Successful Experience>}
For this given question, you immediately came up with a similar success case in memory as follows:

\textbf{[Database schema]}
\begin{verbatim}
CREATE TABLE address_status (
    "address_status" TEXT COMMENT address_status; VALUES: [Active,Inactive],
    "status_id" INTEGER COMMENT status_id; VALUES: [1,2]
    PRIMARY KEY ("status_id")
);

CREATE TABLE customer_address (
    "status_id" INTEGER COMMENT status_id; VALUES: [1,2],
    "address_id" INTEGER COMMENT address_id; VALUES: [606,266],
    "customer_id" INTEGER COMMENT customer_id; VALUES: [1,2]
    PRIMARY KEY ("address_id, customer_id")
);

CREATE TABLE customer (
    "customer_id" INTEGER COMMENT customer_id; VALUES: [1,2],
    "email" TEXT COMMENT email; VALUES: [upurdy0@cdbaby.com,rvatini1@fema.gov],
    "first_name" TEXT COMMENT first_name; VALUES: [Ursola,Ruthanne],
    "last_name" TEXT COMMENT last_name; VALUES: [Purdy,Vatini]
    PRIMARY KEY ("customer_id")
);

CREATE TABLE order_status (
    "status_id" INTEGER COMMENT status_id; VALUES: [1,2],
    "status_value" TEXT COMMENT status_value; VALUES: [Order Received,Pending Delivery]
    PRIMARY KEY ("status_id")
);
\end{verbatim}

\textbf{[Question]}  
How many of the customer addresses are inactive?

\textbf{[Successful SQL]}
\begin{verbatim}
SELECT count(*) 
FROM customer_address 
INNER JOIN address_status 
ON customer_address.status_id = address_status.status_id 
WHERE address_status.address_status = 'Inactive';
\end{verbatim}

\textbf{<Failed Experience>}
Here is a case where the query wasn't quite correct:
\textbf{[Failed SQL]}
\begin{verbatim}
SELECT 
    (COUNT(CASE WHEN e.product = 'Credit card' THEN 1 END) / COUNT(*)) * 1.0 AS percentage
FROM 
    client c 
LEFT JOIN 
    events e ON c.client_id = e.client_id 
WHERE 
    c.sex = 'Male';
\end{verbatim}

==========

\textbf{<<<<<NEW CASE>>>>>}

\textbf{[Database schema]}
\begin{verbatim}
CREATE TABLE cards (
    id INTEGER,
    power TEXT,
    uuid TEXT,
    artist TEXT,
    asciiName TEXT,
    availability TEXT
);

CREATE TABLE foreign_data (
    id INTEGER,
    language TEXT,
    uuid TEXT,
    flavorText TEXT,
    multiverseid TEXT,
    name TEXT
);
\end{verbatim}

\textbf{[Question]}
What percentage of cards without power are in French?

\textbf{[NEW SQL]}
\begin{verbatim}
SELECT 
    (COUNT(CASE WHEN f.language = 'French' THEN 1 END) / COUNT(*)) * 100 AS percentage
FROM 
    cards c 
INNER JOIN 
    foreign_data f ON c.uuid = f.uuid 
WHERE 
    (c.power IS NULL OR c.power = '*');
\end{verbatim}
\endgroup
\end{dialogbox}

\end{document}